\documentclass[lettersize,journal]{IEEEtran}
\usepackage{amsmath,amsfonts}
\usepackage{algorithmic}
\usepackage{algorithm}
\usepackage{array}
\usepackage[caption=false,font=footnotesize]{subfig} 
\usepackage{textcomp}
\usepackage{stfloats}
\usepackage{url}
\usepackage{verbatim}
\usepackage{graphicx}
\usepackage{amsmath}
\usepackage{mathptmx}
\usepackage{mathrsfs}
\usepackage{booktabs} 
\usepackage{cite}

\DeclareMathAlphabet{\mathcal}{OMS}{cmsy}{m}{n}
\newcommand{\triangleq}{\mathrel{\stackrel{\triangle}{=}}}
\begin{document}

\title{Joint Task Offloading, Inference Optimization and UAV Trajectory Planning for Generative AI Empowered Intelligent Transportation Digital Twin}

	\author{Xiaohuan Li, \textit{Member, IEEE}, Junchuan Fan, Bingqi Zhang, Rong Yu, \textit{Member, IEEE}, \\ Xumin Huang, and Qian Chen

	\thanks{Xiaohuan Li and Junchuan Fan are with the Guangxi University Key Laboratory of Intelligent Networking and Scenario System (School of Information and Communication, Guilin University of Electronic Technology), Guilin 541004, China (e-mail: lxhguet@guet.edu.cn; fanjunchuan@mails.guet.edu.cn).}
	
	\thanks{Bingqi Zhang is with the Research Institute of Highway Ministry of Transport, Beijing 100000, China (e-malis: bq.zhang@rioh.cn).}
	
	\thanks{Rong Yu is with the School of Automation, Guangdong University of Technology, Guangzhou 510006, China (e-mail: yurong@gdut.edu.cn).}
	
	\thanks{Xumin Huang is with the School of Automation, Guangdong University of Technology, Guangzhou 510006, China, and also with the Guangxi University Key Laboratory of Intelligent Networking and Scenario System (School of Information and Communication, Guilin University of Electronic Technology), Guilin 541004, China (e-mail: huangxu\_min@163.com).}
	
	\thanks{Qian Chen is with the School of Architecture and Transportation Engineering, Guilin University of Electronic Technology, Guilin 541004, China (e-mail: chenqian@mails.guet.edu.cn).}

}


\IEEEpubid{}

\maketitle

\begin{abstract}
	
To implement the intelligent transportation digital twin (ITDT), unmanned aerial vehicles (UAVs) are scheduled to process the sensing data from the roadside sensors. At this time, generative artificial intelligence (GAI) technologies such as diffusion models are deployed on the UAVs to transform the raw sensing data into the high-quality and valuable. Therefore, we propose the GAI-empowered ITDT. The dynamic processing of a set of diffusion model inference (DMI) tasks on the UAVs with dynamic mobility simultaneously influences the DT updating fidelity and delay. In this paper, we investigate a joint optimization problem of DMI task offloading, inference optimization and UAV trajectory planning as the system utility maximization (SUM) problem to address the fidelity-delay tradeoff for the GAI-empowered ITDT. To seek a solution to the problem under the network dynamics, we model the SUM problem as the heterogeneous-agent Markov decision process, and propose the sequential update-based heterogeneous-agent twin delayed deep deterministic policy gradient (SU-HATD3) algorithm, which can quickly learn a near-optimal solution. Numerical results demonstrate that compared with several baseline algorithms, the proposed algorithm has great advantages in improving the system utility and convergence rate.

\end{abstract}

\begin{IEEEkeywords}
Intelligent transportation digital twin; unmanned aerial vehicles; diffusion model; deep reinforcement learning.
\end{IEEEkeywords}

\section{Introduction}
\IEEEPARstart{I}{ntelligent} transportation digital twin (ITDT) exploits the digital twins (DTs) technology to create DT models of the physical entities by capturing the real-time sensing data, and employ the DT models to facilitate the simulation, understanding, analysis and control of the entire intelligent transportation system (ITS) \cite{1}. The ITDT promises to significantly achieve the full life-cycle network monitoring, accurate transportation safety warning, and informed transportation decision-making for the ITS. To fulfill the ITDT, a variety of enabling technologies such as 5G/6G cellular communications, unmanned aerial vehicles (UAVs), artificial intelligence (AI) and mobile cloud/ edge computing have been proposed for coping with the massive data streams and promoting the network intelligence \cite{57}. As one of the advanced AI technologies, generative artificial intelligence (GAI) can automatically generate, manipulate, and modify different sensing data to address the issues of data scarcity, noise, and biases in high-dynamic environment. We are motivated to employ the GAI to provide high-quality and valuable sensing data and make the updated DTs precisely reflect the physical entities \cite{7}\cite{67}. With the integration of GAI with ITDT, we propose a new networking paradigm termed by GAI-empowered ITDT, which facilitates the high-fidelity DT creation for ITDT.
%which enables the intelligent data synthesis, high-fidelity DT creation, and enhanced system resilience for the ITDT.

The implementation of the proposed GAI-empowered ITDT consists of three stages: raw sensing data collection, GAI-enabled data processing, and high-fidelity DT updating. In the first stage, a number of roadside sensors are deployed on the roads to periodically sense the physical entities of the ITS \cite{vv1}. In the second stage, the sensing data collected by roadside sensors may be missing or abnormal. The state-of-the-art GAI models (e.g., diffusion model) are employed to process the raw sensing data, e.g., performing the data augmentation \cite{62}, imputing the missing data \cite{63}, and repairing the anomalous data \cite{64}. Thus, a series of computation-intensive GAI-enabled data processing tasks are generated on the roadside sensors. In this paper, we also call the above computation tasks of the roadside sensors as the diffusion model inference (DMI) tasks. In addition, UAV have the advantages of ubiquity, high mobility, and low cost, and are leveraged as aerial computing nodes to provide computation offloading and enable the resource-constrained roadside senors to offload the DMI tasks, saving the time and energy consumption. Finally, reliable and accurate state information is obtained for the high-fidelity ITDT updating, which aims to reduce the deviation between the DT models and physical entities.

Nevertheless, there are still challenges that needs to be addressed for the large-scale applications of GAI-empowered ITDT. First, since the UAVs with the limited computing resources are dynamically scheduled to compute the DMI task from geo-distributed roadside sensors. Thus, both the task offloading and the UAV trajectories should be well addressed to let different UAVs successfully serve more roadside sensors within the same time periods. 
Second, for the ITDT updating, we necessitate to tackle the tradeoff between the ITDT fidelity and ITDT updating delay. %In this paper, we pay attention to the diffusion model inference (DMI) task. 
Regarding the DMI, prior studies have demonstrated that the output quality of a pre-trained diffusion model depends on the number of inference steps\cite{10}. Performing more inference steps is beneficial to generate the high-quality data that is important for the ITDT fidelity; nevertheless, it will result in more computational workloads and prolong the ITDT updating delay. To ensure the real-time virtual-real synchronization while satisfying the ITDT fidelity requirement, we could perform the joint optimization of both the DMI task and the UAV sides for the proposed GAI-empowered ITDT. As a result, there exists a challenging joint optimization problem of task offloading, inference optimization, and resource allocation.
Last but not least, seeking a solution to the joint optimization problem in the dynamic and large-scale complex environment is nontrivial. Compared with the conventional convex optimization approaches, multi-agent deep reinforcement learning (MADRL) algorithm is superior to solving the complex decision-making tasks under the network dynamics. MADRL algorithm enables multiple agents to collaboratively seeking a near-optimal solution while making the solution robust to the changes in the network dynamics. However, the current MADRL algorithms should be revised to cope with the challenges such as actor network update conflict and optimization direction deviation in the heterogeneous agents \cite{65}. %so as to enhance the learning performance in terms of the learning accuracy and convergence rate.

To address the above challenges, we study a novel system utility maximization (SUM) problem for the GAI-empowered ITDT, and propose an improved MADRL algorithm to approximate near-optimal optimization policies and makes quick decisions to adapt to the changing environment. In the system model, a set of UAVs are dynamically scheduled to fly and provide computation offloading for the nearby roadside sensors while computing the DMI tasks. %We also optimize the inference steps the diffusion models to process the collected sensing data. 
To fulfill the GAI-empowered ITDT in a comprehensive manner, we aim to maximize the long-term system utility relevant with the overall ITDT fidelity and ITDT updating delay. In the studied SUM problem, we jointly design the task offloading strategy, the inference optimization of all diffusion models, and the UAV trajectory. We also consider feasible constraints such as computing resource capacity, the ITDT fidelity and delay requirements.

Recent works have demonstrated that DRL enables and enhances the multi-agent decision-making in complex and dynamic environments \cite{66}. To obtain an adaptive solution for the heterogeneous-agent optimization problem, we convert the original formulation into a heterogeneous-agent Markov decision process (MDP) that can be addressed by the Twin Delayed Deep Deterministic Policy Gradient (TD3). Inspired by the work \cite{13}, we further revise the TD3 algorithm by using the sequential update mechanism. This replaces the conventional simultaneous updates with the sequential update mechanism that leads to more stable actor network improvement. Specifically, the sensor agent actor networks are updated first and the UAV agent actor networks are subsequently updated based on the latest sensor agent actor networks. The approach can avoid the actor networks conflicts and improving the stability of heterogeneous-agent learning. We are motivated to adopt the sequential update-based Heterogeneous-agent TD3 (SU-HATD3) algorithm to efficiently learn a near-optimal cooperative strategy for the proposed SUM problem. The detailed contributions of this work are listed as follows.

\begin{itemize}

	\item{We present the system model of the GAI-empowered ITDT. In this system model, each roadside sensor generates a DMI task for ITDT update. The UAVs are dynamically scheduled to computer the computationally intensive DMI tasks offloaded from the roadside sensors, which involve processing the sensing data to enhance the ITDT. The cloud server aggregates the processed sensing data from all UAVs and uses this data to update the high-fidelity ITDT.}
	
	\item{We study the SUM problem to maximize the long-term system utility while satisfying necessary ITDTs constraints. To balance the tradeoff between the ITDT fidelity and ITDT updating delay, we study how to offlaod the DMI tasks to UAVs. Meanwhile, we investigate the inference optimization of diffusion models and UAV trajectory planning based on the offloading decisions to perform the ITDT updates.}
	
	\item{We propose the SU-HATD3 algorithm to solve the SUM problem. The algorithm uses the sequential update mechanism. The proposed algorithm addresses the strong cooperation and coupling among heterogeneous agents, ensuring the monotonic improvement of their actor network through a sequential update mechanism. Compared with the baseline DRL algorithms the proposed algorithm has great advantages in improving the long-term system utility and achieving the faster convergence rate.}

		%in both the convergence rate and system utility.}
		
\end{itemize}

The structure of the paper is outlined below. Section II provides a review of related works. Section III introduces the system model. In Section IV, we present the problem formulation. A SU-HATD3 algorithm is proposed in Section V. Section VI is dedicated to the simulation experiments and their analysis, while Section VII provides the conclusion.

\section{Related Works}\label{}

In this section, we review the related work on the GAI-empowered ITDT, UAV scheduling for DT updating and performance comparison.

\subsection{GAI-empowered ITDT}  
Extensive research has been conducted on the application of DT in ITS. In ITS, DT serves as a virtual representation of physical entities to simulate their attributes, behaviors, and interactions\cite{v1}. In \cite{r2}, the authors proposed the DT-based transportation signal control framework in ITS, where sensors provide real-time trajectories and transportation flows for modeling, prediction and simulation. In \cite{r3}, the authors defined DT as a module that reconstructs detailed digital road models to simulate scenarios. In \cite{r4}, a DT framework was developed for ITS by integrating real-time traffic data with SUMO simulation to support transportation management and safety decisions.

Despite challenges such as complex modeling, low simulation efficiency and deviations in practical applications, the emergence of GAI has provided DT with a new foundation of efficiency and accuracy\cite{r6}. In \cite{r7}, a GAI-enhanced DT framework was proposed using federated and MARL for real-time awareness and resource optimization in ITS. In \cite{r8}, GAI was employed for data augmentation to improve DT-based traffic simulation and adaptive modeling in ITS. In \cite{r9}, a GAI-driven DT framework with mechanisms such as ‘Priority Pooling’ and ‘Twin Adapter’ was designed to enhance Internet of Things (IoT) modeling and adaptive management. In \cite{r10}, a GAI-based flow model was embedded into urban DT to improve the modeling and prediction of urban flows, supporting sustainable smart city development. 

\subsection{UAV Scheduling for DT Updating}
UAVs have the advantages of high mobility, easy deployment and good line-of-sight link, which can provide aerial computing to realize data collection and DT updating with low delay, low energy consumption and high precision\cite{r12}. In previous UAV scheduling for DT updating research, optimization objectives such as low delay or high-fidelity are typically achieved by adjusting UAV trajectories, bandwidth allocation, or task offloading ratios \cite{r13,2,42,r16}. In \cite{r13}, UAV swarms were employed as distributed nodes, integrating federated learning and aerial computing with heterogeneous sensing and devices scheduling to improve DT accuracy and energy efficiency. In \cite{2}, a semantic communication framework optimized the UAV CPU frequency and transmission power to improve the timeliness and energy efficiency of DT synchronization. In \cite{r16}, UAV-based DT construction leveraged parameter identification and data assimilation to optimize the model variables, significantly enhancing the DT fidelity and simulation performance. In \cite{42}, a hierarchical digital twin network was constructed through heterogeneous UAV clusters, where cluster-head selection and Digital Twin synchronization optimization improved communication efficiency and reduced transmission delay.

\subsection{Performance Comparison}
Existing studies have demonstrated that GAI model can enhance DT performance. However, most of those works overlook the fidelity and delay requirements of DT updating. Due to the limited computing capability of roadside sensors, UAVs are essential for computing DMI tasks and flexibly updating the DTs, enabling efficient and dynamic DT updating. Nevertheless, the integration of GAI-empowered ITDT with UAV has not been thoroughly explored in existing research. 

Building on this motivation, we consider the SUM optimization problem for the GAI-empowered ITDT with UAV, with the objective of maximizing the ITDT fidelity while minimizing the total update delay. Unlike traditional computation tasks that are either fully completed or dropped, the DMI task can be partially computed with varying computing resource, leading to different output qualities. To accommodate this unique characteristic, we jointly optimize the task offloading, inference optimization and UAV trajectory planning under dynamic environments and limited UAV computing resources. 

Furthermore, while some studies utilize MARL to address UAV scheduling, they often encounter challenges such as training instability and slow convergence, particularly in large-scale heterogeneous systems remain significant limitations. In contrast, our work introduces an enhanced MARL-based framework that supports cooperative decision-making between the UAVs and sensors, achieving both fast convergence and higher system utility in GAI-empowered ITDT.

\begin{figure}
	\centerline{\includegraphics[width=3.5in]{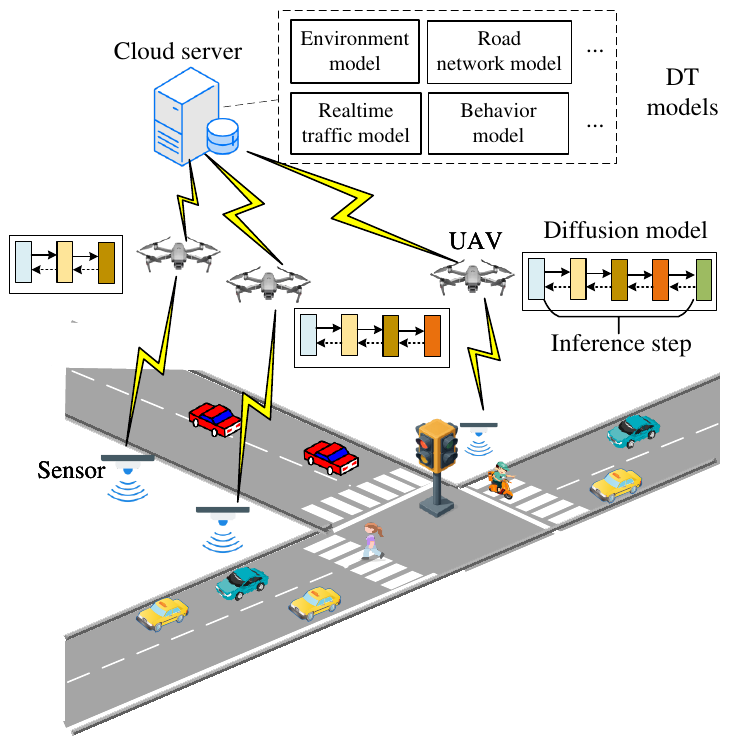}}
	\caption{System model. \label{fig1}}
\end{figure}

\section{SYSTEM MODEL}
As illustrated in Fig. \ref{fig1}, we present the system model of the proposed GAI-empowered ITDT. There are several key entities to support the model functionalities, i.e., roadside sensor, UAV, and cloud server. The following provides additional details about the network entities.

\begin{itemize}
	\item Roadside sensor: A variety of roadside sensors such as roadside LiDAR and camera are deployed to collect the sensing data as the state information of the physical entities of the ITS, e.g., vehicle and pedestrian. However, the raw sensing data is always sparse and low-quality in the inherently complex and dynamic traffic environment, which suffers from the missing values or noise contamination. The noisy or incomplete sensing data necessitates to be processed before the utilization. To this end, diffusion models are exploited to process the sensing data, causing a computation-intensive DMI task on each roadside sensor. Particularly, computation workloads of a DMI task is controllable by adjusting the number of inference steps on demand.
	\item UAV: The DMI task computations brings the rather large computational burdens to the resource-constrained roadside sensors. Diffusion models are deployed on the UAVs to process the DMI tasks, such as performing data augmentation, imputing missing values, and repairing anomalies. To accelerate the DMI task processing for the real-time ITDT updating, a set of UAVs are dynamically scheduled to fly to provide computation offloading for the roadside sensors when necessary. In this paper, we consider the binary offloading scenario between the roadside sensors and UAVs.
	\item Cloud server: After completing the DMI tasks, high-quality sensing data is submitted by the UAVs to the cloud server. The cloud server updates the DT models based on these reliable and accurate data sources, such as environment model, road network model, realtime traffic model, and behavior model. We also consider that the cloud server acts as a centralized manager to solve the studied joint optimization problem of task offloading, inference optimization, and UAV trajectory planning in a centralized manner.
\end{itemize}

\section{PROBLEM FORMULATION}
To implement the GAI-empowered ITDT, $R$ roadside sensors and $N$ UAVs are deployed for the sensing data collection and processing. Let $\mathcal{R}$ and $\mathcal{N}$ represent the sensor set and the UAV set, respectively. Each sensor and UAV is indexed by $r$ and $n$, respectively. The system time horizon $\mathcal{T}$ is divided into $T$ time slots, where each slot $t$ has a duration of $\tau =\mathcal{T}/T$. For simplicity, all roadside sensors are randomly distributed in a rectangular area with the size $l \times l$ and each UAV maintains a steady altitude $H$ in the rectangular area. Each roadside sensor periodically collects real-time sensing data from the corresponding physical entity and a diffusion model is employed to improve the quality of the collected sensing data. At time slot $t$, roadside sensor $r$ collects $s_{r,t}$ amounts of sensing data and generates a DMI task for the sensing data processing. Thanks to the UAV-assisted computation offloading, the DMI task can be entirely offloaded to a proper UAV when the UAV approaches to the roadside sensor. We consider that each DMI task is indivisible and offloaded to a UAV and each UAV computes the DMI tasks at each time slot. The binary offloading decision set of all roadside sensors is $\mathcal{A}=\left\{ {{\alpha }_{n,r,t}}|n\in \mathcal{N},r\in \mathcal{R},t\in \mathcal{T} \right\}$, where $\alpha_{n,r,t}=1$ means that roadside sensors $r$ offloads the DMI task to UAV $n$ at time slot $t$, otherwise the offloading decision is not chosen.

\subsection{Diffusion Model Inference}

We take the DMI task offloading between roadside sensors $r$ and UAV $n$ as an example and let $\alpha_{n,r,t}=1$. Then UAV $n$ receives the sensing data of roadside sensor $r$, and allocates a portion of computing resource $S_{n,r,t}$ to compute the DMI task with the computational frequency $f_{n,r,t}$. Here, we define $S_{n,r,t}$ denotes the computing resource allocated by UAV $n$ for DMI task $r$, which corresponds to the number of inference step and its unit is steps. We define $f_{n,r,t}$ as the computation frequency of the UAV to compute the DMI task and its unit is steps/s. Note that the output quality of the diffusion model depends on the number of the inference steps, and determines the sensing data quality. %The overall quality of all sensing data from different roadside sensors has a critical impact on the the DT fidelity when updating the DT of the ITS based on the sensing data.
Previous studies \cite{18}\cite{27} show that the number of inference steps plays a positive impact on the output quality of a diffusion model. Note that during the diffusion model inference, performing more inference steps (i.e., denoising steps) leads to the higher output quality. Utilizing the sensing data with the higher quality is helpful to improve the ITDT fidelity. In the case $\alpha_{n,r,t}=1$, we evaluate the ITDT fidelity gain contributed by processing the sensing data of roadside sensor $r$ as
\begin{equation}
	F_{{r,t}}=
	\begin{cases}
		\frac{F_{{r}}^{\mathrm{max}}}{S_{{r}}^{\max}-S_{{r}}^{\min}}\left(S_{n,r,t}-S_{{r}}^{\min}\right), \ \ \ \ \ \ S_{{r}}^{\min}\leq S_{n,r,t}\leq S_{{r}}^{\max}\\
		0, \ \ \ \ \ \ \ \ \  \ \ \ \ \ \ \ \ \ \ \ \ \ \ \ \ \ \ \ \ \ \ S_{n,{r},t}< S_{{r}}^{\max} \\
		{F_{{r}}^{\mathrm{\max}}}, \ \ \ \ \ \ \ \ \ \ \ \ \ \ \ \ \ \ \  \ \ \ \ \ \ \ \ S_{n,{r},t}> S_{{r}}^{\max}&
	\end{cases}
	\label{eq11}
\end{equation}
where the $S_{r}^{\min}$ and the $S_{r}^{\max}$ are the minimal and maximal inference steps, respectively, and the $F_r^{\max}$ is the maximal DT fidelity gain by means of inferring the entire diffusion model with sufficient steps.

\subsection{UAV-Assisted Computation Offloading}
Next, we introduce the UAV mobility model and DMI task offloading model in the UAV-assisted computation offloading.

1) \textit{UAV mobility model:} Without the loss of generality, we consider a three-dimensional Cartesian coordinate system where the horizontal position of each roadside sensor at time slot $t$ is known as ${q}_{{{r,t}}}=\left[x_{r,t}, y_{r,t}, 0\right]$ and the instantaneous horizontal position of UAV $n$ at time slot $t$ is given by $q_{n,t}=\left[x_{n,t},y_{n,t}, H\right]$. The UAV position is changed over time according to the given flight angle $\theta_{n,t}$ and a constant velocity $v_{n,t}$. Thus, we update the position of UAV $n$ by
\begin{align}
	x_{n,t+1}=x_{n,t}+\nu_{n,t}\tau\cos(\theta_{n,t})\\
	y_{n,t+1}=y_{n,t}+\nu_{n,t}\tau\sin(\theta_{n,t})
	\label{eq2}
\end{align}

Besides, there exist basic constrains for the UAV mobility. The flight distance constraint of UAV $n$ is expressed by
\begin{align}
	 \parallel q_{n,t+1}-q_{n,t} \parallel \leq L_{\max}, \forall t
	\label{eq3}
\end{align}
where $ L_{\max}$ is the maximal flight distance of the UAV. The flight speed constraint of UAV $n$ is expressed by
\begin{align}
	v_{\min}\le v_{n,t}\le v_{\max}, \forall t
	\label{eq4}
\end{align}
where $v_{\min}$ and $v_{\max}$ are the minimal and maximal speed of the UAV, respectively. The flight area constraint of UAV $n$ is expressed by
\begin{align}
	0\le x_{n,t} \le l, 	0\le y_{n,t} \le l, \forall t
	\label{eq5}
\end{align}
%\noindent where \eqref{eq3} constrain the maximum flight distance of the UAV during each time slot, $\parallel\cdot\parallel$ represents the Euclidean distance, \eqref{eq3} represent the speed constraint, and \eqref{eq4} and \eqref{eq5} constrain the motion range of the UAV.

2) \textit{DMI task offloading model:} If $\alpha_{n,{r},t}=1$, roadside sensor $r$ submits the sensing data to UAV $n$ for the DMI task offloading. Similar to the work in \cite{28}, we consider a time-invariant channel, where the statistical characteristics of the channel remain constant over extended periods. The free-space path loss is assumed in the communication channel between roadside sensor $r$ and UAV $n$ and we calculate the channel gain by
\begin{equation}
	h_{n,r,t}=h_{n}^0 d_{n,{r},t}^{-2}
	\label{eq7}
\end{equation}
where $d_{n,r,t}=\parallel q_{n,t}-q_{r,t}\parallel$ is the communication distance, $h_n^0$ is the channel gain at one meter reference distance, and the path loss exponent is set as 2. Furthermore, we adopt the orthogonal frequency division multiple access technology to eliminate the interference among the roadside sensors. We calculate the achievable data transmission rate between roadside sensor $r$ and UAV $n$ by 
\begin{equation}
	R_{n,{r},t}=\log\left(1+\frac{P_{{r}}\left\|h_{n,r,t}\right\|^{2}}{\sigma_{0}^{2}}\right)
	\label{eq8}
\end{equation}
where $P_{r}$ is the transmit power of the roadside sensor and $\sigma_{0}^{2}$ is the noise power. After that, the sensing data transmission delay between roadside sensor $r$ and UAV $n$ is 
\begin{equation}
	D_{n,{r},t}^{\text{trans}}=\alpha_{n,{r},t}\frac{s_{{r,t}}}{R_{n,{r},t}}
	\label{eq9}
\end{equation}.

For the GAI-enabled sensing data processing, the DMI task is performed on UAV $n$. Based on the computing resources $S_{n,r,t}$ allocated to the DMI tasks, the inference delay of the diffusion model by UAV $n$ is expressed as

\begin{equation}
	D_{n,r,t}^{\text{infer}}=\frac{S_{n,{r},t}}{f_{n,{r},t}}
	\label{eq10}
\end{equation}

Finally, we utilize $\alpha_{n,{r},t}(D_{n,r,t}^{\text{infer}}+D_{n,{r},t}^{\text{trans}})$ to express the DMI task completion delay, which includes both sensing data transmission delay and DMI task inference delay. 

\subsection{SUM Problem Formulation}
We define the system utility at time slot $t$ as the ITDT fidelity gain of processing all sensing data by using the diffusion models minuses the total time consumption for the ITDT updating. On one hand, the ITDT fidelity gain of processing all sensing data is evaluated by $\sum\nolimits_{1\leq r \leq R}{F_{r,t}}$. On the other hand, we neglect the output data transmission delay on the UAVs that help upload the output data of the DMI tasks to the cloud server for the ITDT updating, and roughly measure the total time consumption for the ITDT updating by  
\begin{equation}
	D_{r,t}=\sum\limits_{r=1}^{R}(D_{n,{r},t}^{\mathrm{infer}}+D_{n,{r},t}^{\mathrm{trans}}), \forall n\in{\mathcal{N}}
	\label{eq12}
\end{equation}

Then the system utility at time slot $t$ is expressed by
\begin{equation}
	{{U}_{t}}=\sum\limits_{r=1}^{R}{\left( {{\varphi }_{1}}{{F}_{r,t}}-{{\varphi }_{2}}{{D}_{r,t}} \right)}
	\label{eq13}
\end{equation}
where $\varphi_1$ and $\varphi_2$ are two presetting weighting coefficients. 

In this paper, we aims to maximize the long-term system utility of the GAI-empowered ITDT subject to feasible constraints. We jointly optimize the task offloading, the inference steps of all diffusion models, and the UAV trajectory. To achieve maximize system utility, three key elements are optimized: 1) The DMI task offloading, denoted as $\mathcal{A}=\left\{ {{\alpha }_{n,r,t}},\forall n\in \mathcal{N},\forall r\in \mathcal{R}, t\in \mathcal{T} \right\}$; 2) The inference optimization of all diffusion models, denoted as ${\cal S} = \left\{ {{S_{n,r,t}},\forall n \in {\cal N},\forall r \in {\cal R}},t\in \mathcal{T} \right\}$; 3) The UAV trajectory planning, denoted as $\mathcal{G}=\left\{ \theta _{n,t}^{{}},v_{n,t}^{{}},\forall n\in \mathcal{N}, t\in \mathcal{T} \right\}$. Accordingly, we express the long-term SUM problem with necessary constraints as follows
\begin{equation}
	\begin{aligned}
		& P:\text{   }\underset{\mathcal{A},\mathcal{S},\mathcal{G}}{\mathop{\max }}\,\underset{T\to \infty }{\mathop{\lim }}\,\frac{1}{T}\sum\limits_{t=1}^{T}{{U}_{t}} \\ 
		& s.t.\text{    }C1: \sum_{n=1}^{N}\alpha_{n,{r},t}\leq1,\forall n\in{\mathcal{N}},\forall r\in{\mathcal{R}},\forall T\in{\mathcal{T}}\\
		& \ \ \   \text{    }C2: 0\leq F_r^{min}\leq F_{{r,t}}, \forall r\in{\mathcal{R}},\\
		& \ \ \   \text{    }C3: \sum_{r=1}^{R}\alpha_{n,{r},t}{{{S}_{n,{{m}_{r}},t}}\le {{S}_{n}}},\forall n\in{\mathcal{N}},\forall r\in{\mathcal{R}},\forall t\in{\mathcal{T}}\\
		& \ \ \   \text{    }C4: D_{r,t}\leq D_{r}^{\max}<\tau,\forall r\in{\mathcal{R}}, \forall t\in{\mathcal{T}}\\
		& \ \ \   \text{    }C5: (2)-(6) \\ 
	\end{aligned}
	\label{eq14}
\end{equation}

Constraint C1 indicates each DMI task can only offload to one UAV in each time slot. Constraint C2 present the ITDT fidelity gain requirements. Constraint C3 represents the resource constraint of the UAV. Constraint C4 represents the GAI task completion delay requirement. Constraint C5 defines the mobility limitation of UAV. The problem $P$ has the following properties that make it challenging to solve. First, the problem is a nonconvex mixed-integer nonlinear programming problem. Specifically, the problem has the integer and continuous decision variables, thus exhibits the non-convex properties. Second, the solution should be adaptive to quickly react to the dynamic environment. Furthermore, considering the formulated problem exhibits characteristics of the long-term accumulation, we can transform this problem into a multi-agent decision-making problem and employ the MADRL approach to address it.

%\noindent where $C1$ represents the DMI task of roadside sensor can only be offloaded to one UAV, $C2$ the DT minimum tolerated fidelity, $F_r^{min}$ represents the minimum fidelity tolerance, $C3$ represents the computing resource constraint of the UAV, $C4$ denotes the DT maximum tolerated delay, $D_{r}^{\max}$ represents the maximum delay tolerance, $C5$ represents the UAV trajectory constraint.

\section{THE PROPOSED SU-HATD3 ALGORITHM}

\begin{figure*}[htbp]
	\centerline{\includegraphics[scale=1.0]{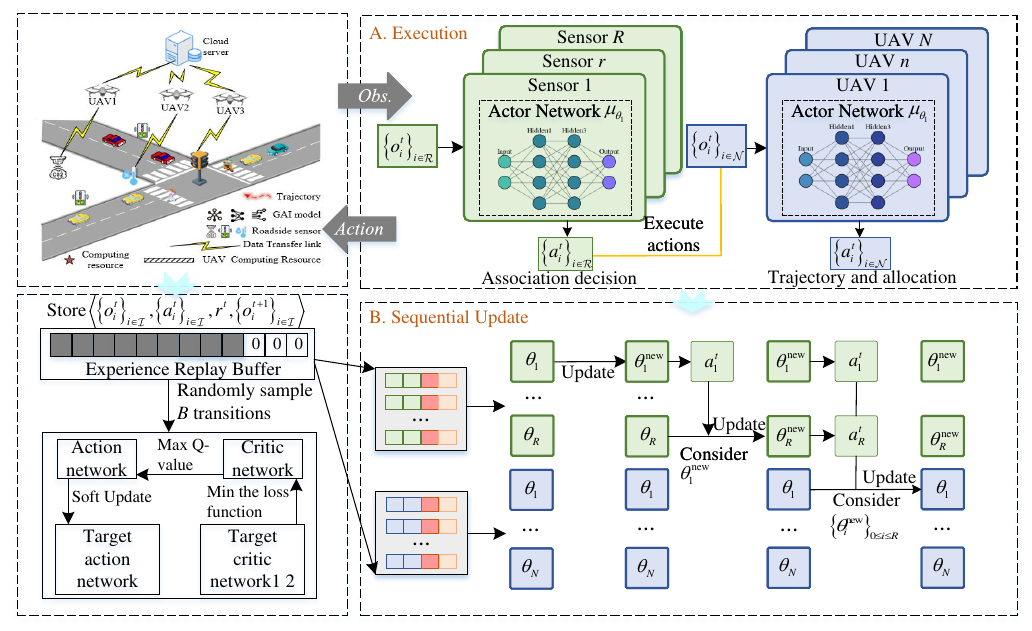}}
	\caption{The diagram of the proposed SU-HATD3 algorithm.\label{fig3}}
\end{figure*}

In this section, we integrate the sequential update mechanism into the commonly utilized MADRL algorithm, i.e., MATD3 algorithm, and present a SU-MATTD3 algorithm to efficiently address the problem $P$ by capturing the relationships between the sensor states and the UAV states. We introduce the the motivation of SU-HATD3, the heterogeneous-agent MDP, the SU-HATD3 algorithm architecture and the computational complexity analysis. 

\subsection{Motivation of SU-HATD3 Algorithm}

The motivation for the sequential update mechanism is to enhance optimization efficiency and system utility in the heterogeneous agents environment. The traditional parallel updates often lead to policy conflicts and unstable convergence \cite{40}\cite{53}. In this approach, sensor agents first determine the task offloading strategy, followed by UAV agents allocating computing resources and planning their trajectories. The sequential update mechanism addresses this by ensuring that each agents decisions are based on the most recent updates from other agents, reducing the likelihood of conflicting policies. This ordered update process enhances coordination among agents, accelerates convergence, and ultimately improves system utility.

\subsection{Heterogeneous-agent MDP Model}

To alleviate the complexity of decision-making and find near-optimal solutions, we define sensor agents and UAV agents. The set of agents is denoted as $i \in \mathcal{I} \triangleq \{1, 2, \ldots,R,R+1,\cdots, R+N\}$. Thus, the problem is formulated as the heterogeneous-agent MDP with $R$ sensor agents and $N$ UAV agents. In particular, the interaction process for sensor agents and UAV agents is characterized by the tuple $\left\langle \left\{{o}_{i,t}\right\}_{i\in \mathcal{I}},\left\{{a}_{i,t}\right\}_{i\in \mathcal{I}},{r}_t,\left\{{o}_{i,{t+1}}\right\}_{i\in \mathcal{I}}\right\rangle$. The $\left\{ {o}_{r,t}\right\}_{r \in \mathcal{R}} $ and $\left\{ {o}_{n,t}\right\}_{n \in \mathcal{N}} $ define the observation spaces for sensor agents and UAV agents, respectively. The $\left\{ {a}_{i,t}\right\}_{i \in \mathcal{R}} $ and $\left\{ {a}_{i,t}\right\}_{i \in \mathcal{N}} $ represent the action sets available to sensor agents and UAV agents, respectively. The $r_t$ denotes the reward function. The observation space, action space, and reward function for the sensors and UAVs are described for each time slot $t$.

\textit{1) MDP of Sensor Agents}

The sensor agent mainly selects the UAVs for the DMI task offloading. Aiming at determining the offloading strategy, sensor agents require observe the maximum inference step, their own locations, the computing resource of UAV, as well as the UAVs' locations.

\textit{ a) Observation}: At time slot $t$, the observation of sensor agents $r$ is expressed as
\begin{equation}
	{o}_{r,t} =\left\{ \left. S_{r}^{max},q_{r,t},q_{n,t},T_n,\forall n \in \mathcal{N}  \right\}, r \in \mathcal{R} \right.
	\label{eq15}
\end{equation}

It is worth noting that each sensor can only access its own location, while the coordinates of other sensors remain unknown. In contrast, the positions of all UAVs are known to the sensors because UAVs are broadcast their locations.

\textit{ b) Action}: At time slot $t$, the observation of sensor agents $r$ is expressed as

\begin{equation}
	{a}_{r,t} =\left\{ \left. \alpha_{n,{r},t}
	,\forall n \in \mathcal{N}  \right\}, r \in \mathcal{R} \right.
	\label{eq16}
\end{equation}

\textit{2) MDP of UAV Agents}

After the sensor agents determines the offloading strategy, the UAV agents makes the inference optimization and trajectory decision according to the observed offloaded the sensor information. 

\textit{ a) Observation}: At time slot $t$, the observation of UAV agents is expressed as

\begin{equation}
	{o}_{n,t} =\left\{ \left. q_{n,t},q_{r,t},S_{r}^{max}, F_{r}^{max},\forall r \in \mathcal{R}_{n}'  \right\}, n \in \mathcal{N} \right.
	\label{eq18}
\end{equation}

\noindent where $R_n'$ represents the set of sensors that offload DMI tasks to the UAV $n$.

\textit{ b) Action}: At time slot $t$, the observation of UAV agents is expressed as

\begin{equation}
	{a}_{n,t} =\left\{ \left. \theta _{n,t},v_{n,t},S_{n,{r},t} ,\forall r \in \mathcal{R}_{n}' 
	\right\}, n \in \mathcal{N} \right.
	\label{eq19}
\end{equation}

\noindent where $\theta _{n,t}\in[-1,1]$ represents the flight angle of the UAV, $v_{n,t}\in[-1,1]$ represents the flight speed of the UAV, $T_{n,r,t}\in[-1,1]$ represents an unnormalized weight indicating the relative proportion of computing resources to be allocated for the DMI task $r$. Since the total available computing resource of UAV is limited, the actual allocated computing resource for the DMI task is computed via normalization as

\begin{equation}
	\tilde{S}_{n,r,t}=\frac{S_{n,r,t}}{\sum_{r^{\prime}\in\mathcal{R}_n^{\prime}}S_{n,{r},t}} S_n,\forall r \in R_{i}'
	\label{eq20}
\end{equation}

\noindent where $\tilde{S}_{n,r,t}$ is the final number of inference steps assigned to model.

\textit{ c) Reward}: Each agent receives an immediate reward from the environment after performing an action at time slot 
$t$. The system utility is considered as the immediate reward. If a DMI task fails to meet its completion deadline or if the ITDT update fails to achieve the required fidelity, a penalty term $\rho_{r,t}$ will be further incorporated into the reward to impose a punishment. Therefore, the long-term discounted cumulative reward across $T$ time slots can be represented as

\begin{equation}
	{{r}_{t}}=U- \sum_{r=1}^R\rho_{r,t}
	\label{eq17}
\end{equation}

\noindent weher $\rho_{r,t}$ can be represented as

\begin{equation}
	\rho_{r,t}=max(D_{r,t}-D_{r}^{\max},0)+max(F_r^{min}-F_{r,t},0)
	\label{eq21}
\end{equation}

\subsection{The SU-HATD3 Algorithm Architecture}

To address the above heterogeneous-agent MDP, considering the cooperation among heterogeneous agents for ITDT updates, the SU-HATD3 algorithm is proposed. As shown in Fig. \ref{fig3}, the sensor agents and UAV agents input their respective observations into the corresponding actor networks to determine the optimal actions for the current time step. After executing actions in the environment, the agents receive new observations and immediate rewards. The tuples $\left\langle \left\{{o}_{i,t}\right\}_{i\in \mathcal{I}},\left\{{a}_{i,t}\right\}_{i\in \mathcal{I}},{r}_{t},\left\{{o}_{i,t+1}\right\}_{i\in \mathcal{I}}\right\rangle$ are stored in the cloud server's experience replay buffer. The cloud server uses these experiences for centralized training to update the actor network and global critic networks, and applies a soft update mechanism to maintain training stability. The updated actor networks are distributed to the agents, enabling them to make more efficient and coordinated decisions in the next round of environment interaction.

In the SU-HATD3 framework, sensor agents and UAV agents are assigned actor networks $\left\{ {{\mu }_{{{\theta }_{i}}}} \right\}_{i\in \mathcal{I}}$ and target actor networks $\left\{ {{\mu }_{{{\theta }_{i}}'}} \right\}_{i\in \mathcal{I}}$. Each actor network ${{\mu }_{{{\theta }_{i}}}}$ generates the action based on local observation. To evaluate the joint actions of all agents, two critic networks ${{\left\{ {{Q}_{{{\phi }_{k}}}} \right\}}_{k=1,2}}$ and two target critic networks ${{\left\{ {{Q}_{{{\phi }_{k}}'}} \right\}}_{k=1,2}}$ are deployed on the cloud server. These critic networks estimate the global Q-values using the combined observations and actions of all agents, thereby guiding the actor network optimization. At each time step $t$, trajectories are collected and used to update the networks over $T$ time steps (i.e., $T$ time slots). All collected samples are centrally processed on the cloud server. The server aggregates interaction data from all agents and updates the actor networks and critic networks.

\textit{1) Execution phase}

In the execution phase, the heterogeneous agents interact with the environment based on their current actor networks. Due to the strong coupling between sensor and UAV agents, a collaborative decision-making is adopted. For sensor agents, the DMI task offloading decision of sensor agents is discrete. This paper convert the continuous action output of the actor network into a discrete decision by using a discretization mapping. The action of sensor agent $i$ at time slot $t$ is given by

\begin{equation}
	a_{r,t}={{\mu }_{{{\theta }_{r}}}}(o_{r,t})+\mathcal{X}_{t},r \in \mathcal{R}
	\label{eq22}
\end{equation}

\noindent where $o_{i,t}$ denotes the local observation and $\mathcal{X}_t$ represents the exploration noise. After executing action ${{\left\{ a_{r,t} \right\}}_{r \in \mathcal{R}}}$, UAV agent obtains the local observation ${{\left\{ o_{n,t} \right\}}_{n \in \mathcal{N}}}$. The action of UAV agent $n$ at time slot $t$ is expressed as

\begin{equation}
	a_{n,t}={{\mu }_{{{\theta }_{n}}}}(o_{n,t})+\mathcal{X}_{{t}},\text{  }n \in \mathcal{N}
	\label{eq23}
\end{equation}t

After all agents execute their actions, the environment provides the global reward $r_{t}$ and updates the observations for the next time step. The tuples $\left\langle o_{i,t},a_{i,t},r_{i,t},o_{i}^{t+1} \right\rangle$ are stored in the experience replay buffer $\mathcal{B}$ for subsequent training.

\textit{2) Update phase} 

The distinctive observation and action spaces associated with sensor and UAV agents often cause training instability and pose challenges for convergence \cite{46}. To handle the complexity of the MDP involving $R+N$ agents, we adopt the sequential update mechanism. In detail, the multi-agent advantage decomposition lemma suggests that sequential actor networks updates can effectively improve cooperation stability in heterogeneous multi-agent systems, as each agent’s actor networks optimization is conditioned on the latest actions of the others\cite{13}. Consequently, the sequential update mechanism allows each agent to optimize actor network while considering the most recent behaviors of others, thus preventing conflicts in actor updates and ensuring monotonic policy improvement.

In the update phase, the SU-HATD3 algorithm updates all actor–critic networks by using experiences sampled from the experience replay buffer. Specifically, the sensor agents update their actor networks first. The UAV agents then update their actor networks based on the latest sensor agent actor networks. This sequential update design improves coordination among heterogeneous agents and stabilizes convergence. The loss function $L$ is formulated as the mean squared error between the predicted Q-values and target Q-values, which is minimized to optimize the critic networks ${{Q}{{{\phi }{1}}}}$ and ${{Q}{{{\phi }{2}}}}$.

\begin{multline}
	L\left( {{\phi }_{k}} \right)={{\mathbb{E}}_{\left( {{o}},{{a}},{{r}},{{o}^{'}} \right)\sim \mathcal{B} \text{}}}\left[ {{\left( {{y}_{t}}-{{Q}_{{{\phi }_{k}}}}\left( \left\{ o_{i,t} \right\}_{i\in \mathcal{I}},\left\{ a_{i,t} \right\}_{i\in \mathcal{I}} \right) \right)}^{2}} \right],\\ k=1,2
	\label{eq24}
\end{multline}

\noindent where the target Q-values ${{y}_{t}}$ is computed as:

\begin{multline}
	{{y}_{t}}={{r}_{t}}+\gamma \min {{Q}_{{{\phi }_{k}}'}}\left( \left\{ o_{i,t+1} \right\}_{i\in \mathcal{I}},\left\{ \hat{a}_{i,t+1} \right\}_{i\in \mathcal{I}} \right),k=1,2
	\label{eq25}
\end{multline}

\noindent where $\left\{ \hat{a}_{i,t+1} \right\}_{i=1}^{\mathcal{I}}$ is generated from $\left\{ {{\mu }_{{{\theta }_{i}}'}} \right\}_{i=1}^{\mathcal{I}}$.

After updating the critic networks, the algorithm proceeds to sequentially update the actor networks. The sensor agent $r$ actor network are preferentially updated as

\begin{align}
	& \theta _{r}^{new}= \notag\\
	&\arg \underset{{{\theta }_{r}}}{\mathop{\max }}\, 
	{{\mathbb{E}}}\Bigg[
	{{Q}_{{{\phi }_{1}}}}\Big(
	\big\{ o_{i,t} \big\}_{i\in \mathcal{I}},
	{{\mu }_{{{\theta }_{r}}}}\big( \{ o_{r,t} \}_{r\in \mathcal{R}} \big),
	{{\mu }_{{{\theta }_{n}}}}\big( \{ o_{n,t} \}_{n\in \mathcal{N}} \big)
	\Big) 
	\Bigg]
	\label{eq26}
\end{align}

\noindent and the UAV agents $n$ are subsequently updated based on the updated sensor agent $r$ actor network

\begin{align}
	& \theta _{n}^{new} = \notag\\ 
	&\arg \underset{{{\theta }_{n}}}{\mathop{\max }}\,\mathbb{E}\Bigg[
	{{Q}_{{{\phi }_{1}}}}\Big(
	\left\{ o_{i,t} \right\}_{i\in \mathcal{I}},
	\underbrace{{{\mu }_{\theta _{r}^{new}}}\left( \left\{ o_{r,t} \right\}_{R\in \mathcal{R}} \right)}_{\text{Latest sensor networks}},
	{{\mu }_{{{\theta }_{n}}}}\left( \left\{ o_{n,t} \right\}_{n\in \mathcal{N}} \right)
	\Big)
	\Bigg]
	\label{eq27}
\end{align}

Finally, the target networks are softly updated to ensure stability

\begin{equation}
	\begin{gathered}
		{{\phi }_{1}}' \leftarrow \tau {{\phi }_{1}}+(1-\tau ){{{\phi }_{1}}'}, \\ 
		{{\phi }_{2}}' \leftarrow \tau {{\phi }_{2}}+(1-\tau ){{{\phi }_{2}}'}, \\ 
		{{\theta }_{i}}' \leftarrow \tau {{\theta }_{i}}+(1-\tau ){{{\theta }_{i}}' }
	\end{gathered}
	\label{eq28}
\end{equation}

Algorithm~\ref{alg:alg1} presents the pseudocode of the proposed SU-HATD3 algorithm for solving the system utility maximization problem. In the execution phase, the agents interact with the environment to sequentially make decisions (Lines 3–18). Each sensor agent observes $o_{i,t},i\in \mathcal{R}$ and makes an association decision $a_{i,t},i\in \mathcal{R}$ (Lines 6–9). UAV agents perform inference optimization for the DMI task and plan the trajectories based on the observation $o_{i,t},i\in \mathcal{N}$ (Lines 11–14). The environment returns new states and rewards. These transitions are stored in the experience replay buffer (Lines 16–17). In the update phase, a batch of transitions is sampled from the experience replay buffer (Line 20). The critic networks are updated (Line 21). Then, the actor networks are updated sequentially. The sensor agent actor networks are updated based on the current UAV agent actor networks (Line 23). The UAV agents then update their actor networks based on the last sensor agent actor networks (Lines 26). This sequential update ensures that each agent actor network is optimized individually. This also maintains coordination between the sensor agents and the UAV agents.

\begin{algorithm}[t]
	\caption{SU-HATD3 Algorithm}
	\label{alg:alg1}
	\begin{algorithmic}[1]
		
		\STATE \textbf{Initialize:} actor networks $\{\mu_{\theta_i}\}_{i\in\mathcal{I}}$, 
		target actors $\{\mu_{\theta'_i}\}_{i\in\mathcal{I}}$, 
		critics $Q_{\phi_1},Q_{\phi_2}$, 
		target critics $Q_{\phi_{1}'},Q_{\phi_{2}'}$, 
		experience replay buffer $\mathcal{B}$, training episodes $Epi$, and step length $Ste$.
		
		\STATE \textbf{for} Episode $=1$ \textbf{to} $Epi$ \textbf{do}
		\STATE \hspace{1em} Initialize the environment.
		
		\STATE \hspace{1em} \textbf{// Execution phase}
		\STATE \hspace{1em} \textbf{for} step $=1$ \textbf{to} $Ste$ \textbf{do}
		\STATE \hspace{2em} \textbf{for} $r=1$ \textbf{to} $\mathcal{R}$ \textbf{do}
		\STATE \hspace{3em} Sensor agent $r$ acquires observations $o_i^t$ from the environment.
		\STATE \hspace{3em} Sensor agent $r$ takes actions $a_{r,t}$ according to (\ref{eq22}).
		\STATE \hspace{2em} \textbf{end}
		
		\STATE \hspace{2em} Roadside sensors execute actions $\{a_{r,t}\}_{r\in\mathcal{R}}$.
		
		\STATE \hspace{2em} \textbf{for} $i=1$ \textbf{to} $\mathcal{N}$ \textbf{do}
		\STATE \hspace{3em} UAV agent $n$ acquires observations $o_i^t$ from the environment.
		\STATE \hspace{3em} UAV agent $n$ takes actions $a_{n,t}$ according to (\ref{eq23}).
		\STATE \hspace{2em} \textbf{end}
		
		\STATE \hspace{2em} Execute actions $\{a_{n,t}\}_{n\in\mathcal{N}}$ in the environment.
		
		\STATE \hspace{2em} Observe next observations $\{o_{i,{t+1}}\}_{i\in\mathcal{I}}$ and reward $r^t$.
		
		\STATE \hspace{2em} Push transition $\langle o_{i,t}, a_{i,t}, r_t, o_{i,t+1} \rangle$ into $\mathcal{B}$.
		\STATE \hspace{1em} \textbf{end}
		
		\STATE \hspace{1em} \textbf{// Update phase}
		\STATE \hspace{1em} Sample a batch of $b$ transitions from $\mathcal{B}$.
		
		\STATE \hspace{1em} Update the global critic networks with parameters $\phi^1,\phi^2$ by formula (\ref{eq24}).
		
		\STATE \hspace{1em} \textbf{for} $r = 1$ \textbf{to} $\mathcal{R}$ \textbf{do}
		\STATE \hspace{2em} Update sensor agent $r$ actor network according to (\ref{eq26}).
		\STATE \hspace{1em} \textbf{end}
		
		\STATE \hspace{1em} \textbf{for} $n = 1$ \textbf{to} $\mathcal{N}$ \textbf{do}
		\STATE \hspace{2em} Update UAV agent $n$ actor network according to (\ref{eq27}).
		\STATE \hspace{1em} \textbf{end}
		
		\STATE \hspace{1em} Smoothly update the target critic networks utilizing (\ref{eq28}).
		
		\STATE \textbf{end}
		\begin{flushright}
			
		\end{flushright}		
	\end{algorithmic}
\end{algorithm}

\subsection{Computational Complexity Analysis}

The computational complexity of the SU-HATD3 algorithm can be expressed as $\mathcal{O}\left( E\left[ \left( R+N \right)B{{L}_{a}}+2B{{L}_{c}} \right] \right)$, where $E$ denotes the number of training episodes, $R$ and $N$ represent the numbers of sensor agents and UAV agents, respectively, $B$ is the batch size. The term ${{L}_{a}}$ and ${{L}_{c}}$ represent the sizes of the actor network and critic networks, respectively, and are related to factors such as the number of layers and neurons per layer. The complexity is dominated by the need to update both the actor network and critic networks. Specifically, $\mathcal{O}\left( \left( R+N \right)B{{L}_{a}} \right)$ accounts for the sequential optimization of the actor networks, which scales linearly with the number of agents and the size of each actor network. The $\mathcal{O}\left( B{{L}_{c}} \right)$ reflects the computational cost of updating the global critic networks. Therefore, the overall complexity depends on several factors, including the number of agents, the size of the neural networks, and the minibatch size.

\begin{figure*}[!t]
	\centering
	\subfloat[The reward under different algorithms.]{
		\includegraphics[width=0.31\textwidth]{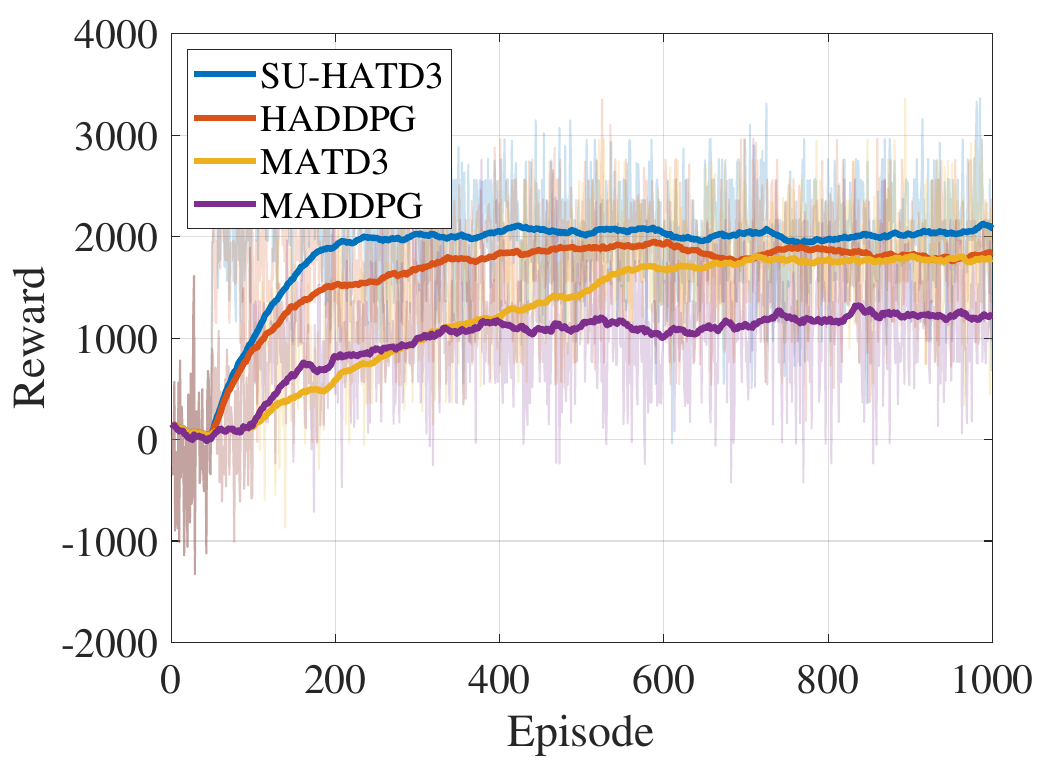}
		\label{fig4a}
	}
	\hfill
	\subfloat[The reward under different learning rates.]{
		\includegraphics[width=0.31\textwidth]{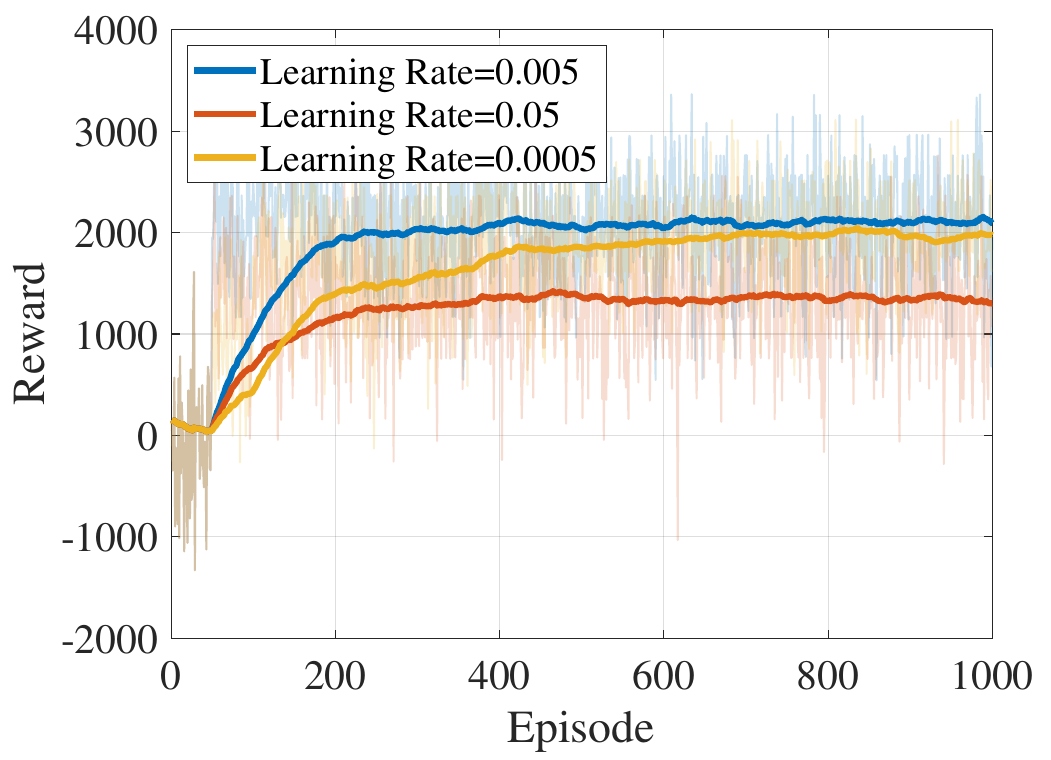}
		\label{fig4b}
	}
	\hfill
	\subfloat[The reward under different scenarios scale.]{
		\includegraphics[width=0.31\textwidth]{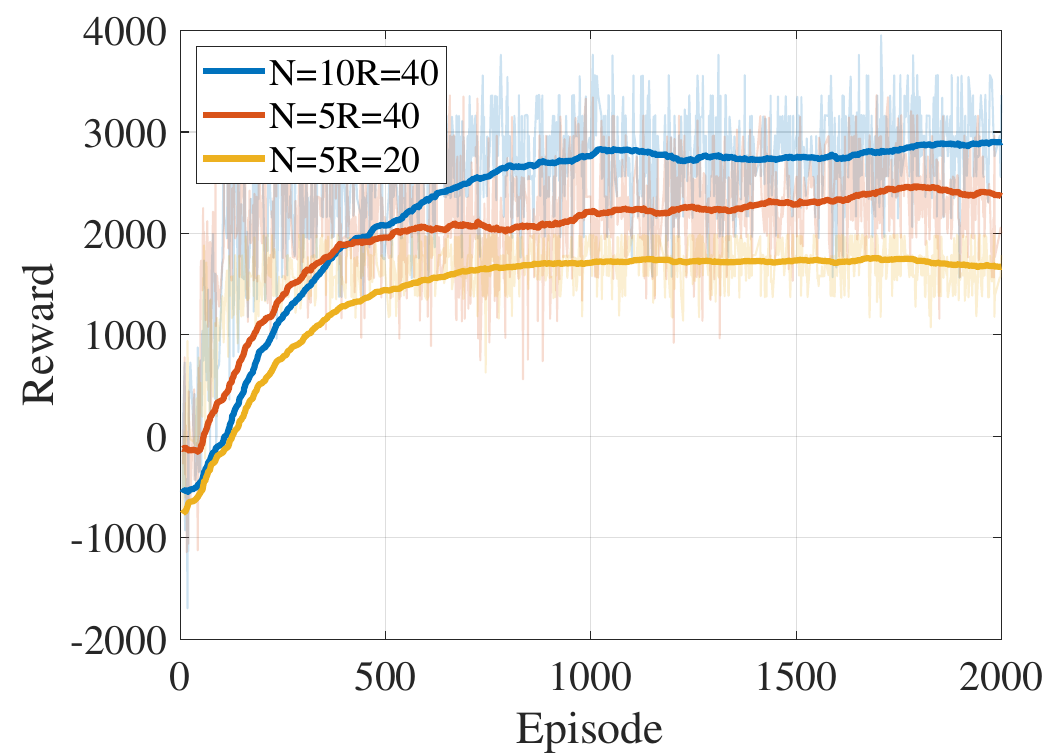}
		\label{fig4c}
	}
	\caption{Convergence analysis under different algorithms, learning rates, and scenario scales.}
	\label{fig4}
\end{figure*}

\begin{figure*}[!t]
	\centering
	\subfloat[The system utility under different number of UAVs.]{%
		\includegraphics[width=0.3\textwidth]{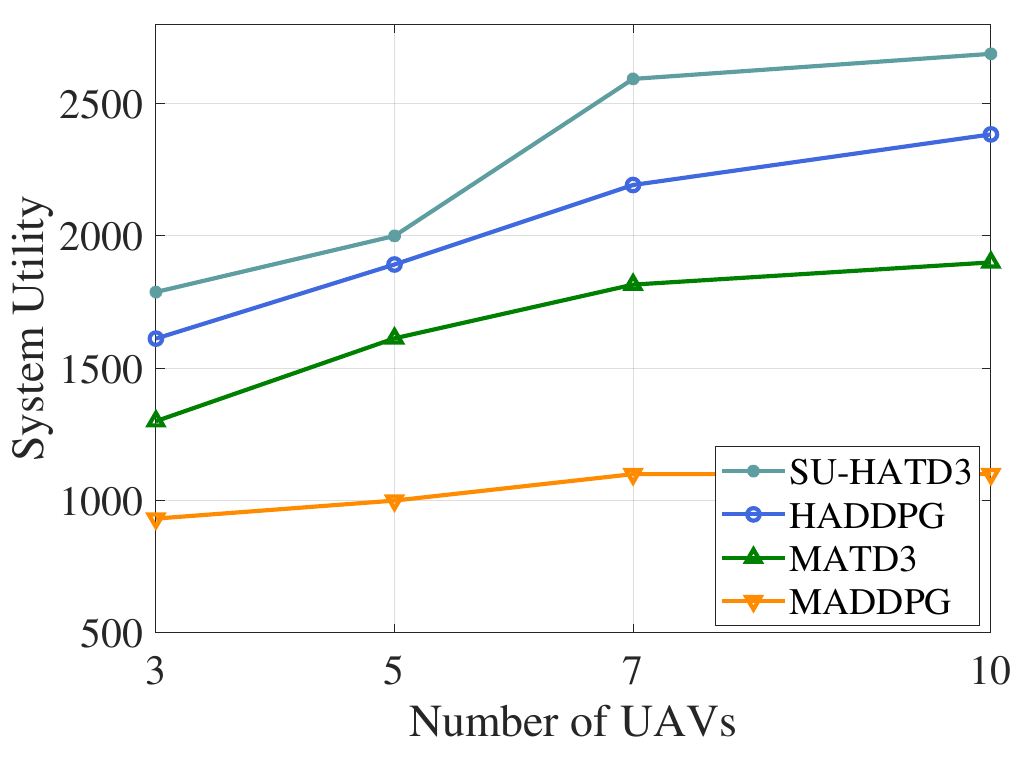}%
		\label{fig6a}}
	\hfill
	\subfloat[The system utility under different number of roadside sensors.]{%
		\includegraphics[width=0.3\textwidth]{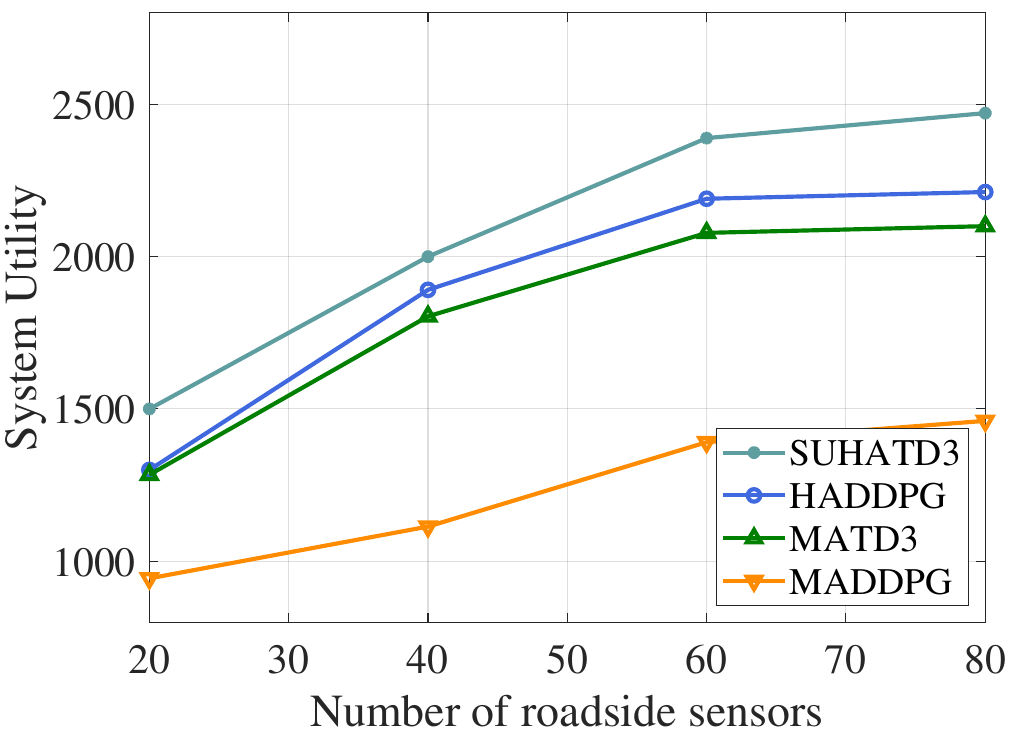}%
		\label{fig6b}}
	\hfill
	\subfloat[The system utility under different computation frequency of UAVs.]{%
		\includegraphics[width=0.3\textwidth]{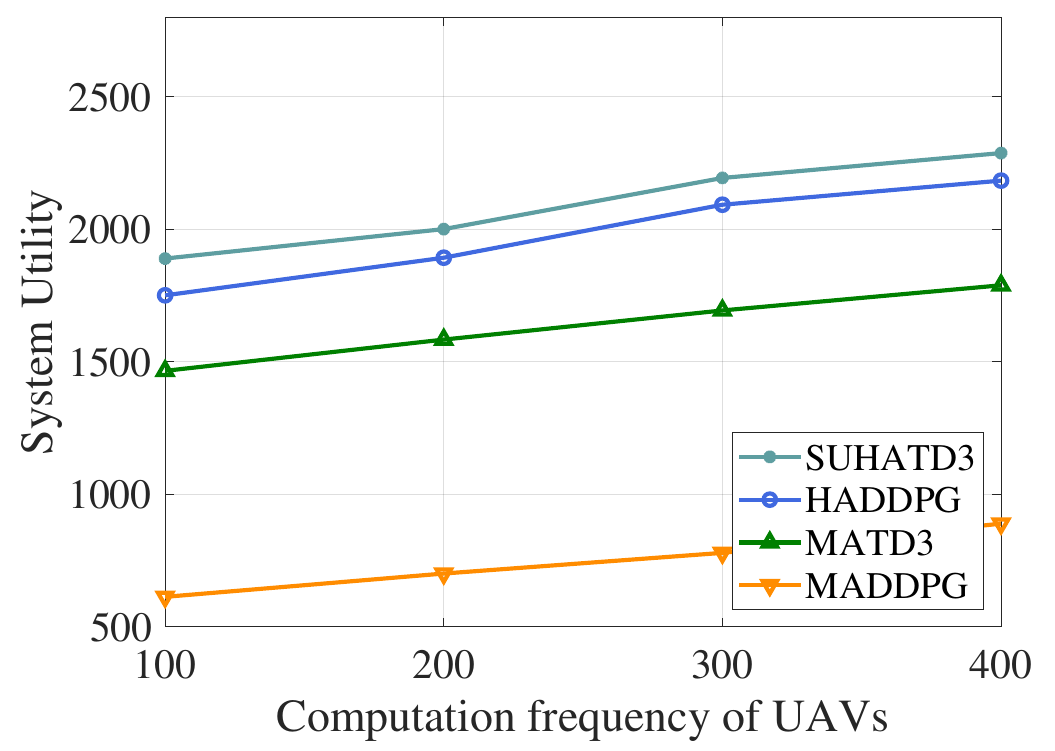}%
		\label{fig6c}}
	\caption{System utility of various algorithms under different settings.}
	\label{fig6}
\end{figure*}

\section{SIMULATION RESULTS}

\subsection{Experiment Setup}

We conduct extensive simulations to evaluate the performance of the SU-HATD3 algorithm in the GAI-empowered ITDT. All experiments are implemented in PyTorch and executed on a workstation equipped with an AMD Ryzen 9 5950X CPU and an NVIDIA RTX 3090 GPU. The same environment is used for all baseline algorithms to ensure consistent experimental conditions. The actor and critic learning rates are set to $1\times {{10}^{-6}}$ and $1\times {{10}^{-3}}$, respectively. We use an experience replay buffer of size ${10}^{6}$. We set the batch size to 256, the discount factor $\gamma $=0.99, and soft-update coefficient $\tau $=0.005. Gaussian exploration noise with a decaying variance is applied during training.

The simulation environment emulates an urban intelligent transportation scenario, where UAVs act as aerial edge servers to execute DMI tasks of the roadside sensors. The scenario area is a $1000 \times1000$ m$^2$ rectangular region, where UAVs are randomly deployed at a fixed altitude of 50 m and roadside sensors are also distributed within the same region. Each roadside sensor collects sensing data and generates a DMI task in every time slot. A complete summary of the parameters and their corresponding values is presented in Table \ref{tab:parameters}. To assess its advantages, a set of benchmark algorithms is also evaluated.

\begin{table}[htbp]
	\centering
	\caption{SIMULATION PARAMETERS}
	\setlength{\tabcolsep}{4pt}
	\begin{tabular}{lll}
		\toprule
		\textbf{Symbol} & \textbf{Definition} & \textbf{Value (Unit)} \\
		\midrule
		$N$ & Number of UAVs & [5,10] \\
		$R$ & Number of roadside sensors & [20,50] \\
		$U$ & Computation frequency of UAVs & [100,400] step/s \\
		$T_n$ & Computing resources of UAVs & $10^6$ step \\
		$v_{n,t}$ & Speed of UAVs & [70, 150] m/s \\
		$s_{r}$ & The amount of sensing data & [0, 24] KB \\
		$S_{r}^{\min}$ & Minimum inference step & 0 \\
		$S_{r}^{\max}$ & Maximum inference step & [400, 600] steps \\
		$F_{{r}}^{\mathrm{min}}$ & Minimum fidelity for DT & 0 \\
		$F_{{r}}^{\mathrm{max}}$ & Maximum fidelity for DT & [50, 150] \\
		$P_{n,{r}}$ & The transmission power of sensor & 0.1 W \\
		$h^0$ & The channel gain & 1 db \\
		\bottomrule
	\end{tabular}
	\label{tab:parameters}
\end{table}

\begin{itemize}
	\item \textbf{MADDPG} is a policy-based reinforcement learning technique that adopts an architecture of centralized training and distributed execution\cite{33}. The algorithm allows each agent to learn its own deterministic policy while utilizing centralized training to acquire global information, thereby promoting coordination among agents in partially observable environments. However, MADDPG may encounter challenges related to training stability and scalability as the number of agents increases.
	
	\item \textbf{MATD3} is an improved version of MADDPG \cite{34}. The algorithm introduces twin Q-values and delayed updates to reduce Q-value estimation bias, improving training stability and scalability. Compared to MADDPG, MATD3 better addresses instability and scalability issues in multi-agent environments.
	
	\item \textbf{HADDPG} is simplified version of SU-HATD3, which uses a single critic network. The algorithm maintains the sequential update mechanism, the absence of double Q-learning limits its performance in handling complex environments. As a result, HADDPG typically shows inferior convergence stability compared to more advanced algorithms like SU-HATD3.
\end{itemize}

\subsection{Performance Evaluation of the SU-HATD3 Algorithm }
\textit{1) Convergence performance}

Fig.~\ref{fig4}\subref{fig4a} compares the performance of different algorithms for solving the optimization problem. Reward aspect: MADDPG has the smallest reward, MATD3 and HADDPG achieve similar rewards. SU-HATD3 achieved the highest reward, outperforming MADDPG, MATD3, and HADDPG by 39\%, 12.56\%, and 11.8\%, respectively. Convergence aspect: MADDPG converges the slowest, followed by MATD3, then HADDPG, with SU-HATD3 achieving the fastest convergence. This demonstrates that SU-HATD3 not only delivers high rewards but also converges more efficiently compared to the baseline algorithms.

In Fig.~\ref{fig4}\subref{fig4b}, the training process of the SU-HATD3 algorithm is demonstrated for different learning rates (0.05, 0.005, and 0.0005). It can be seen from Fig.~\ref{fig4}\subref{fig4b} that the learning rate affects the reward value. A large learning rate may cause the reward to exceed the global optimum too soon, and a small one can lead to slower training and potential stagnation. When the learning rate is small, the convergence speed of computing power becomes slow, which significantly increases the convergence complexity of the network. Simulation results show that the SU-HATD3 algorithm can achieve the best performance with learning rate of 0.005.
	
Fig.~\ref{fig4}\subref{fig4c} verifies the generalization of the SU-HATD3 algorithm in different scale scenarios. Reward aspect: under the same number of UAVs, the reward also increases with the $R$ increase. This shows that in the dense roadside sensors scenarios, the strategy can more fully schedule the computing resources of UAV and prioritize to the roadside sensors that can obtain higher rewards. For the same number of roadside sensors, the system reward increases as the number of UAVs grows. This indicates that with the increase in computing resources, more DMI tasks can be performed. Convergence aspect: as the scene size increases, it also convergence also slows. In summary, the SU-HATD3 algorithm demonstrates good convergence in scenarios of different scales, indicating its strong generalization.

\textit{2) System Utility}

Fig.~\ref{fig6}\subref{fig6a} compares the system utility of four algorithms under different numbers of UAVs. As the number of UAVs rises from 3 to 10, the system utility improves for all algorithms. This is because more UAVs provide additional computing resources, which can be allocated to the DMI task. When the number of UAVs reaches 10, HADDPG achieves a system utility close to 2700. The SU-HATD3 outperforms HADDPG, MATD3, and MADDPG, which achieve system utilities of over 2400, 1750, and 1100, respectively. Overall, our proposed algorithm consistently outperforms the baseline algorithms.

Fig.~\ref{fig6}\subref{fig6b} compares the system utility of four algorithms under different numbers of roadside sensors. As the number of roadside sensors rises from 20 to 80, the system utility improves for all algorithms. This occurs because the increase in DMI tasks provides UAVs with more options. Among these options, the number of low-cost and high-fidelity DMI tasks also increases, which significantly enhances the overall system utility. Overall, our proposed method outperforms the three baseline algorithms in terms of system utility. This advantage mainly stems from the SU-HATD3 algorithm ability to learn effective strategies during training, which further enhances the overall system utility.

Fig.~\ref{fig6}\subref{fig6c} compares the system utility of four algorithms under different UAV computational frequency. The system utility increases as UAV computational frequency improves. This is because higher computational frequency reduce the DMI task completion delay. Specifically, when the UAV computing capability is set to 2000, SU-HATD3 outperforms MADDPG, MATD3, and HADDPG by 150\%, 33\%, and 8\%, respectively.

\begin{figure}
	\centerline{\includegraphics[width=3.5in]{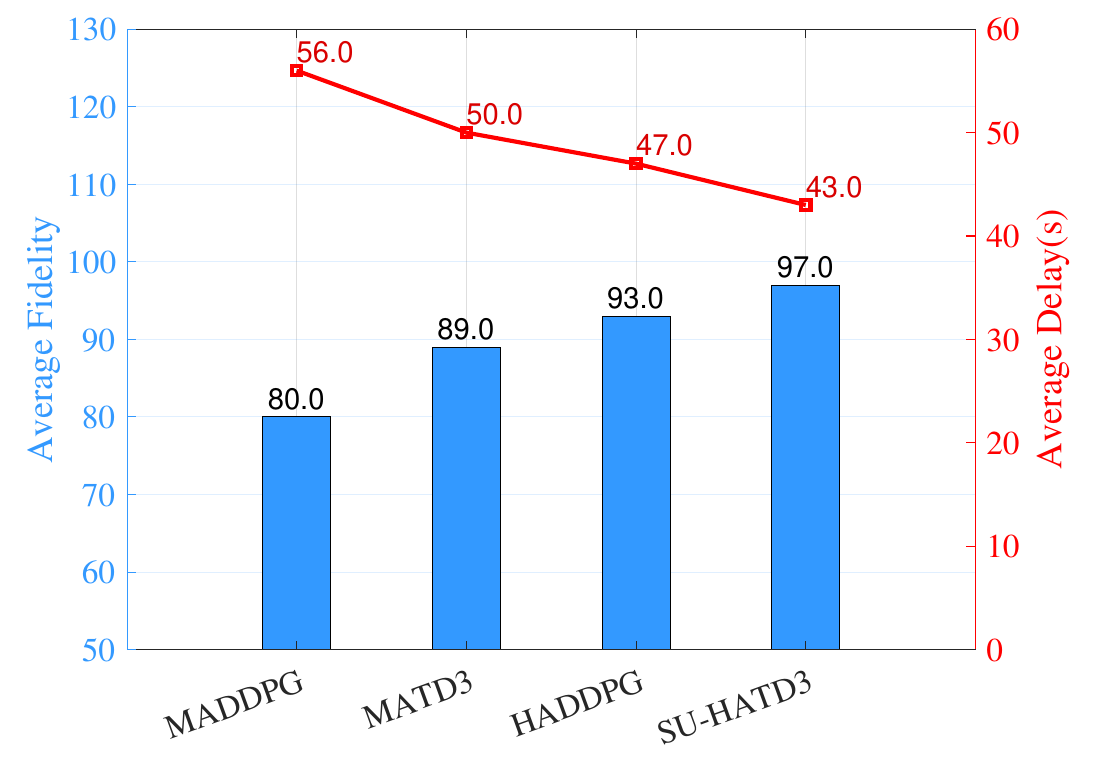}}
	\caption{Comparison of the average fidelity and average delay for 5 UAVs and 40 roadside sensors under different algorithms. \label{fig5}}
\end{figure}

\begin{figure}
	\centerline{\includegraphics[width=3.5in]{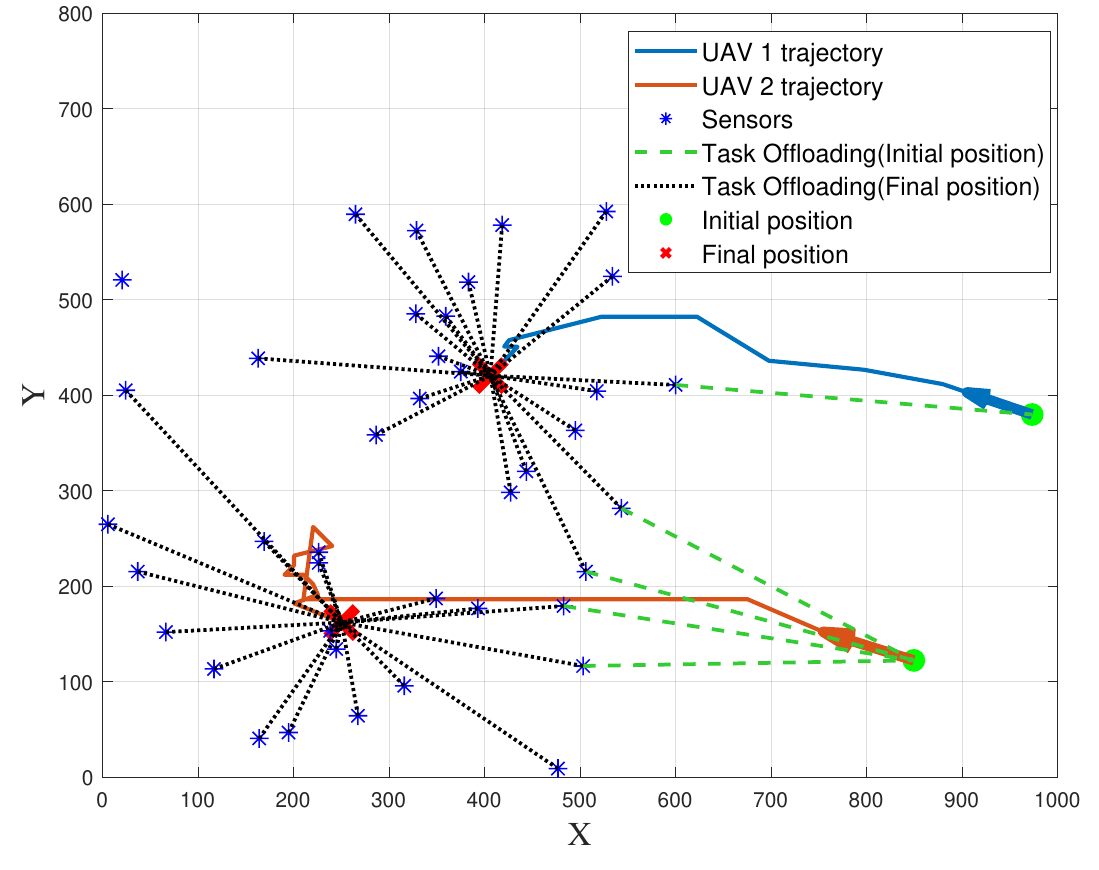}}
	\caption{UAV movement trajectory with $N$ = 2, $R$ = 40 (The start and end points are marked with text and red stars, respectively). \label{fig10}}
\end{figure}

\textit{3) Comparison of different performances}

Fig.~\ref{fig5} shows the performance of different metrics obtained by various algorithms. In terms of fidelity, the SU-HATD3 algorithm is able to update higher fidelity DTs, which is crucial under the computing resource limitations of UAVs. In terms of delay, the SU-HATD3 algorithm can update DT with lower delay, which is essential for addressing the challenge of UAV trajectory planning.

\textit{4)Trajectory Results} 

Fig.~\ref{fig10} evaluates the effectiveness of the SU-HATD3 algorithm by analyzing the UAV trajectories. Fig. \ref{fig10} shows the movement trajectories of 2 UAVs. It can be observed that the initial position of the UAV is away from the roadside sensors, and as the trajectory runs, the UAVs tends to move towards the location regions where multiple roadside sensors are located. This is because the reward function includes the transmission delay term, and being close to the sensor can significantly shorten the communication distance, reduce the transmission delay, and yields a higher system utility. Therefore, the UAV actively adjust its trajectory to move closer to the roadside sensors dense region. In addition, the results show that the UAV has few DMI task offloading when it is far from the sensors. The number of DMI task offloading increases significantly when UAVs is close to the roadside sensors.

\section{CONCLUSION}
To achieve system utility maximization in the GAI-empowered ITDT, we studied a joint optimization problem. In this problem, UAVs execute DMI tasks that are offloaded from the roadside sensors. Each DMI task can be offloaded to an appropriate UAV , which then performs inference optimization of all diffusion models and adjusts its trajectory to improve ITDT fidelity and reduce ITDT update delay. In particular, we consider the fidelity-delay tradeoff for GAI-empowered ITDT. To address the resulting optimization challenge, we developed the SU-HATD3 algorithm. The proposed algorithm tackles the strong cooperation and coupling among heterogeneous agents. The simulation results show that compared to the baseline algorithms, the proposed algorithm has significant advantages in system utility and convergence rate. In future work, we plan to study the combination of three-dimensional trajectory planning and communication resource optimization to enhance the application of UAVs in ITS and better meet the requirements of ITS systems in dynamic and large-scale complex environments

\bibliographystyle{IEEEtran}
\bibliography{main}

\end{document}